\documentclass{article}

\PassOptionsToPackage{numbers, compress}{natbib}

\usepackage[final]{neurips_2022_ml4ps} 

\usepackage[utf8]{inputenc} 
\usepackage[T1]{fontenc}    
\usepackage{hyperref}       
\usepackage{url}            
\usepackage{booktabs}       
\usepackage{amsfonts}       
\usepackage{nicefrac}       
\usepackage{microtype}      
\usepackage{xcolor}         
\usepackage{amsmath}

\usepackage{graphicx}

\title{First principles physics-informed neural network for   quantum wavefunctions and eigenvalue  surfaces}

\author{%
 Marios Mattheakis$^1$\thanks{https://scholar.harvard.edu/marios$\_$matthaiakis}, Gabriel R. Schleder$^1$, Daniel T. Larson$^2$, Efthimios Kaxiras$^{1,2}$ \\
$^1$John A. Paulson School of Engineering and Applied Sciences, Harvard University \\ $^2$Department of Physics, Harvard University
\\
\texttt{ \{mariosmat, gschleder, dtlarson, kaxiras\}@g.harvard.edu},
}

\usepackage[normalem]{ulem}

\newcommand{\rr}{ {\bf r}  }
\newcommand{\RR}{ {\bf R}  }
\newcommand{\HH}{ {\hat {\mathcal {H} } }}
\newcommand{\psiL}{\psi_\text{LCAO}}

\begin{document}

\maketitle

\begin{abstract}
Physics-informed neural networks have been widely applied to learn general parametric solutions of differential equations. Here, we propose a neural network  to discover  parametric eigenvalue and eigenfunction  surfaces of quantum systems. We apply our method to solve the hydrogen molecular ion.  This is an {\em ab initio} deep learning method that  solves the Schr\"odinger equation with the  Coulomb potential  
yielding realistic wavefunctions that 
include a cusp at the ion positions. The neural solutions   are continuous and  differentiable functions of the interatomic distance
and their derivatives are analytically calculated by applying automatic differentiation. Such a parametric and analytical form of the solutions is useful  for further calculations such as the determination of force fields. 
%
\end{abstract}

\section{Introduction}

Physics-informed neural networks (PINNs) have been widely applied recently to study various kinds of differential equations \cite{karniadakisNatureReview2021}. 
Supervised PINNs  can be trained on data to learn nonlinear differential operators \cite{deeONET}, discover differential equations \cite{rudy2017scienceAdv, raissi2019}, and solve inverse problems \cite{mattia2022, inverseOpticsOE2020, alessandro2020}.
%
Unsupervised  PINNs can be trained without using any labeled data to discover analytical and differentiable solutions of ordinary   
 \cite{lagaris1998, mariosHNN} or partial differential equations (PDEs) \cite{spiliopoulos2018,GholamiNips2021}, and eigenvalue problems \cite{Jin2020, fermiNet2020, PRA2021, Jin2022}. 
%
%


In this work we introduce a novel PINN architecture to learn the quantum mechanical wavefunctions for electrons in molecules. This  approach can obtain continuous potential energy surfaces  and the associated parametric wavefunctions.  
In the physics of solids and molecules, the Coulomb potential plays a crucial role since it represents the basic interaction between the system's components, electrons and ions. 
This potential is characterized by singularities which yield cusps in the wavefunction solutions of the Schr\"odinger equation.   Standard numerical approaches consider pseudo-potentials  to avoid this issue
by employing an effective potential of the ion screened by core electrons.
Although this is an efficient approach for numerical calculations, this approximation requires the careful choice and generation of effective potentials to provide consistent wavefunctions and energies; 
it also omits the treatment of core electrons, which in some applications are important. 
Here we show that PINNs can solve the real Coulomb potential in a simple model,
the hydrogen molecular ion  H$^+_2$, thus enabling {\em ab initio} calculations for simple molecules.  
We expect that the method will be generalizable to more complicated cases, which is the subject of ongoing work. 
By exploiting the property of NN's to learn general representations, we obtain continuous and analytically differentiable eigenvalues and wavefunctions that depend on the  interatomic distance, which make 
it feasible to obtain derivatives of the energy landscape;
these derivatives are related to physical 
observables like forces and vibrational 
frequencies.

 {\bf Main contributions}:
    i) We propose a novel PINN architecture for solving the  Schr\"odinger equation for the H$_2^{+}$ ion, including the singular Coulomb potential. Functions including cusps are explicitly included by embedding the approximate Linear Combination of Atomic Orbitals (LCAO) solution. 
    ii) The network computes  continuous and analytically differentiable wavefunction and eigenvalue surfaces for a  range of interatomic distances. Being analytical, these solutions can be used to compute additional physical properties like forces and vibrational frequencies. 
    iii) We obtain  accurate solutions with fast training by embedding  physics-inspired features in the solution like inversion symmetry, and by incorporating the asymptotic behavior through the LCAO approximation.

\section{Related work}
The optimization of PINNs  is achieved by minimizing a loss function constructed by differential equations that encode the physical principles of a system. Specifically,  the loss function may consist of an equation-driven and a data-driven component \cite{raissi2019, karniadakisNatureReview2021, mariosHNN}. The former depends only on the neural solutions and their derivatives with respect to the inputs; the derivatives are calculated using automatic differentiation (autograd). The latter can compare neural predictions to ground truth data \cite{raissi2019} or impose physical laws \cite{mariosHNN}. 

Recently,  PINNs have been employed   to solve quantum eigenvalue problems formulated by the stationary  Schr\"odinger  equation  $\HH \psi = E\psi$, where $\HH$ is a known Hamiltonian operator and  $\psi$, $E$ are, respectively, the unknown  wavefunction and energy we seek to obtain. Typically there are two main deep learning approaches that have been used to solve the Schr\"odinger  equation \cite{Carleo2019,Manzhos2020,Noe2022}.  The method introduced in Refs. \cite{Jin2020, Jin2022} considers a PINN that predicts both  $\psi$ and $E$. By minimizing a loss function to  satisfy the Schr\"odinger PDE, the  NN  approximately calculates  eigen-solutions. 
  The second approach \cite{fermiNet2020, PRA2021} is based on the variational principle. The NN returns only $\psi$, from which the eigen-energy   is computed as the expectation value $\langle {\HH}\rangle =\langle\psi|{\HH}|\psi\rangle/\langle\psi|\psi\rangle$ (in Dirac notation), which  is minimized to optimize the NN. 
  %
%
In this study we adopt the  method used in Ref. \cite{Jin2020, Jin2022}. We generalize this deep learning approach to obtain accurate  generalized parametric eigen-solutions, namely $\psi$ and $E$ as smooth and differentiable functions of the interatomic distance.

\section{{\em Ab initio} deep learning models}

{\bf Hydrogen molecular ion:}
We design a deep NN to obtain the ground state wavefunctions and energies for the single electron in H$^+_2$ as a function of the interatomic distance. For this  we employ a PINN  to solve for the eigenvalues and eigenfunctions of the Hamiltonian operator:
\begin{align}
\label{eq:Hamiltonian}
    \HH = - \frac{1}{2} \nabla^2 -   \frac{1}{| {\rr}-\RR_1|}  -   \frac{1}{| {\rr}-{\RR_2}|},
\end{align}
where we use atomic units, $\rr=(x,y,z)$, and, without loss of generality, the molecule is oriented along the $x$-axis, so $\RR_1 = -\RR_2 = (R,0,0)$.
We seek NN solutions that are continuous functions of $R$, and thus we call them generalized parametric neural solutions.

\begin{figure}
  \centering
  \includegraphics[scale=0.8]{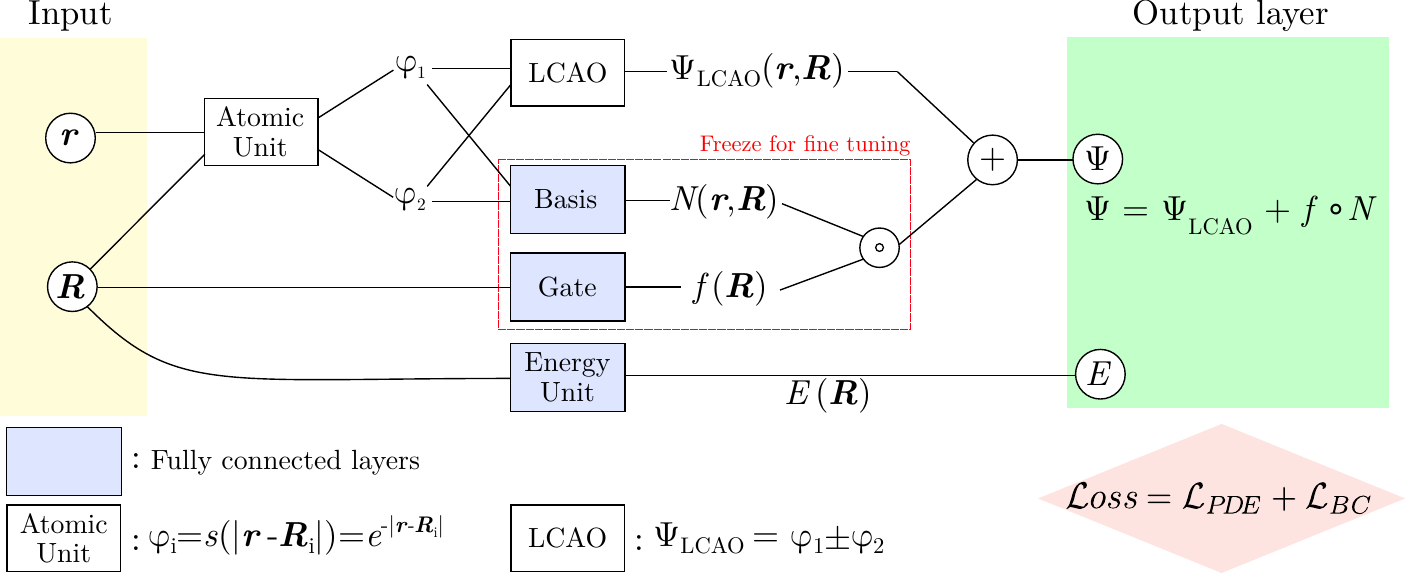}
  \caption{{\em Ab initio} neural network architecture. In the H$_2^+$ configuration, $\boldsymbol{R}$ is a vector of two triples,  $\boldsymbol{R}_i$ is a single vector describing  the position of one nucleus. \label{fig:fig1}}
\end{figure}

{\bf Network architecture:} 
The NN architecture we employ is shown in  Fig. \ref{fig:fig1}. The network inputs are $\rr$, the coordinates of the electron, and the parameter $R$, which determines the geometry of the quantum system.  
As  Fig. \ref{fig:fig1} demonstrates, there are  several units in the architecture and  the forward pass consists of several branches. 
The two inputs 
go through the atomic unit (AU) that returns the hydrogen atomic $s$-orbitals $ \phi_{1,2} = s(|\rr \pm \RR|) = e^{-|\rr \pm \RR|}$ for the left ($x=-R$) and right ($x=R$) ions, respectively. This operation is used  for feature-engineering and  does not contain trainable parameters.
Next, the approximate LCAO solution,  $\psiL=\phi_1 \pm \phi_2$, is constructed by passing $\phi_1$, $\phi_2$ through the LCAO unit;  we do not  provide the normalized $\psiL$ because the neural  $\psi$ will be normalized after the network training. Here we focus on the ground state and use the symmetric $\psiL$ (the ``$+$'' sign); the extension to antisymmetric solutions (the ``$-$'' sign) is straightforward \cite{Jin2022}.   
This approximate solution, which is accurate  as $R\rightarrow\infty$, will be used to build the neural $\psi$.
Though any initial guess is possible, starting from a physics-based asymptotic solution has obvious advantages.
The features $\phi_{1,2}$ also pass through the Basis Unit (BU), which is a multi-layer fully connected feed forward NN (FFNN) that returns a nonlinear combination of $\phi_1$ and $\phi_2$, called $N(\rr,R)$.  This unit respects inversion symmetry by construction \cite{anwesh2022}: $N(x,y,z,R)= \sum_j w_j \left[ B_j(x,y,z,R) + B_j(-x,y,z,R) \right]+b$, where $B_j$ are the outputs of the  neurons in the last hidden layer of the BU, while $w_j$ and $b$ indicates a linear output layer in the BU. 
The purpose of this unit is to correct  $\psiL$ while respecting the inversion symmetry. 
Because $\psiL$ is increasingly more accurate as $R$ increases, we introduce a gate $f(R)$ that is a small FFNN, only depending on $R$, which learns 
the range in $R$ where $N(x,y,z)$ is important.

The full neural wavefunction is thus:
\begin{align}
    \label{eq:psi}
    \psi(\rr,R) = \psiL(\rr,R) + f(R)\circ N(\rr,R),
\end{align}
where $\circ$ indicates the element-wise product.

The neural energy $E(R)$ is the output of a separate FFNN that takes only $R$ as input, and hence is independent of $\rr$ as required from physical considerations
(the total energy of the molecule does not depend on the position of the electrons).  This unit allows the network to learn a smooth and differentiable $E(R)$ that can be evaluated at any $R$.
To improve the accuracy of $E(R)$, we perform a fine-tuning optimization where  we freeze the BU and gate (red box in Fig.~\ref{fig:fig1}) and train the Energy unit (EU) alone. 
Optionally, after the training phase 
we can explicitly calculate $\langle{\HH}\rangle$ using the neural $\psi$ to get a consistent value of the eigen-energy for any desired value of $R$.

{\bf Optimization:} 
The network is optimized by  minimizing a physics-informed loss function:
\begin{align}
    \label{eq:Lpde}
    \mathcal{L} &= \Big\langle \left( \HH \psi(\rr_i, R_i) - E(R_i)\psi(\rr_i,R_i) \right)^2 \Big\rangle_i
        +\  \Big\langle  \psi(\rr_i, R_i) ^2 \Big\rangle_{|\rr_i|>r_\text{cut}},
\end{align}
where the brackets denote averaging over the training points. This loss directs the NN to find $\psi(\rr,R)$ and $E(R)$ that satisfy the Schr\"odinger equation with the Hamiltonian in Eq.~\ref{eq:Hamiltonian} subject to the boundary conditions $\psi(\rr, R)\rightarrow 0$ as $|\rr| \rightarrow\infty$. In practice, the second term in Eq.~\ref{eq:Lpde}, $\mathcal{L}_\text{BC}$, requires that $\psi(\rr,R)$ vanish for any $|\rr|$ larger than a manually selected cutoff value, $r_\mathrm{cut}$.  

The architecture used to solve the H$^+_2$ system consists of 2 hidden layers of 16 neurons each for the BU, one layer of 10 neurons for the gate, and two layers of 32 neurons each for the EU, with a sigmoid activation for all the hidden neurons.  We train the NN using the Adam \cite{adam} optimizer with a learning rate of $ 8\times 10^{-3}$. 
The network is optimized for $5\times 10^3$ epochs but we   save the model with the lowest $\mathcal{L}$.  For the fine-tuning phase we   load the best model and  train only the EU   using the Adam optimizer   with  a learning rate of $10^{-4}$.


\section{Results}

\begin{figure}[ht]
  \centering
\includegraphics[width=.850\textwidth]{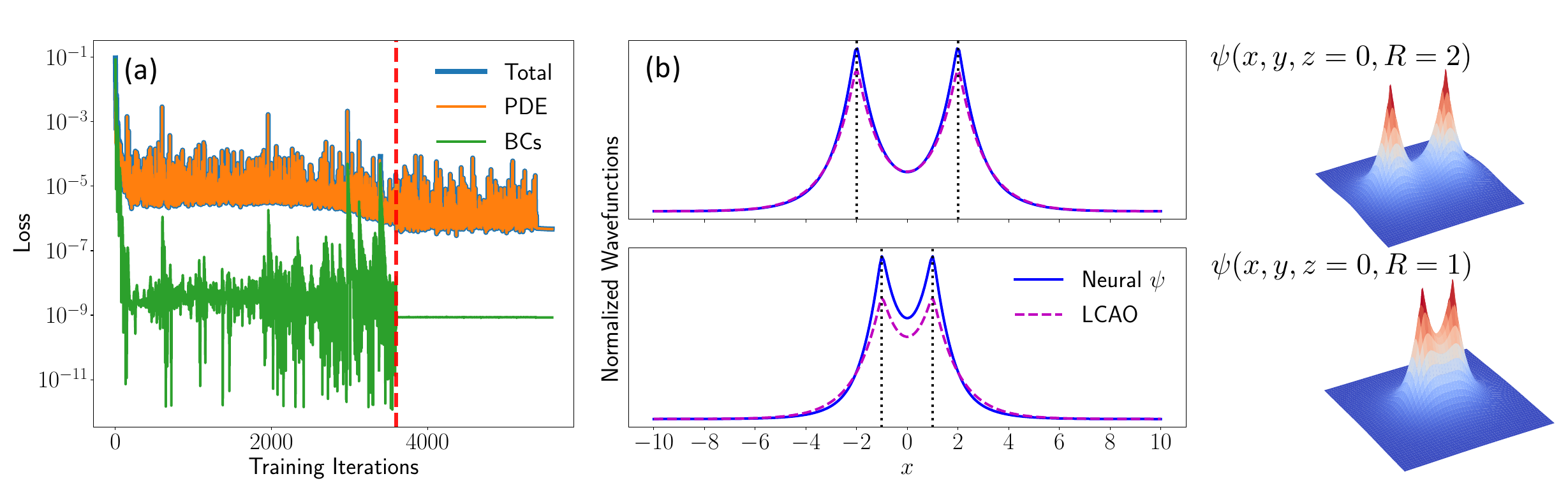}
  \caption{
  (a) The total loss function $\mathcal{L}$, from Eq.\eqref{eq:Lpde}, and its components during the training and fine tuning phases, where the dashed red line separates the two phases. \label{fig:loss}
  (b) Left: Neural $\psi$ (solid blue) and $\psiL$ (dashed red) along the $x$-axis 
  for two different $R$ values, with the ionic positions $\pm R$ indicated by the vertical dashed black lines. Right: $\psi(x,y,0)$ for the same two values of $R$. 
  }
  \label{fig:fig3}
\end{figure}

We employ the PINN desscribed above 
to solve the Schr\"odinger equation for H$^+_2$. The  training set is constructed  by  sampling $10^6$ points in the ranges $(x,y,z) \in [-18,18]$ with a cutoff at $r_\mathrm{cut}=17.5$ and $R \in [0.2,3]$. 
The points are randomly sampled at every epoch  yielding  more robust training \cite{mariosHNN, spiliopoulos2018}. 
In Fig. \ref{fig:fig3}(a) we show the  loss function during the training and fine-tuning phases, where the vertical dashed red line separates the two phases.  Since $\mathcal{L}_\text{BC}$ does not depend on $E$ it does not change  during the fine tuning. The code is written in pytorch \cite{pytorch} and can be found on   github\footnote{\url{https://github.com/mariosmat/PINN_for_quantum_wavefunction_surfaces}}.
The training   takes  less than   5 minutes  on an NVIDIA Tesla V100 GPU with  256 GB memory. 

\begin{figure}[ht]
  \centering
\includegraphics[scale=.18]{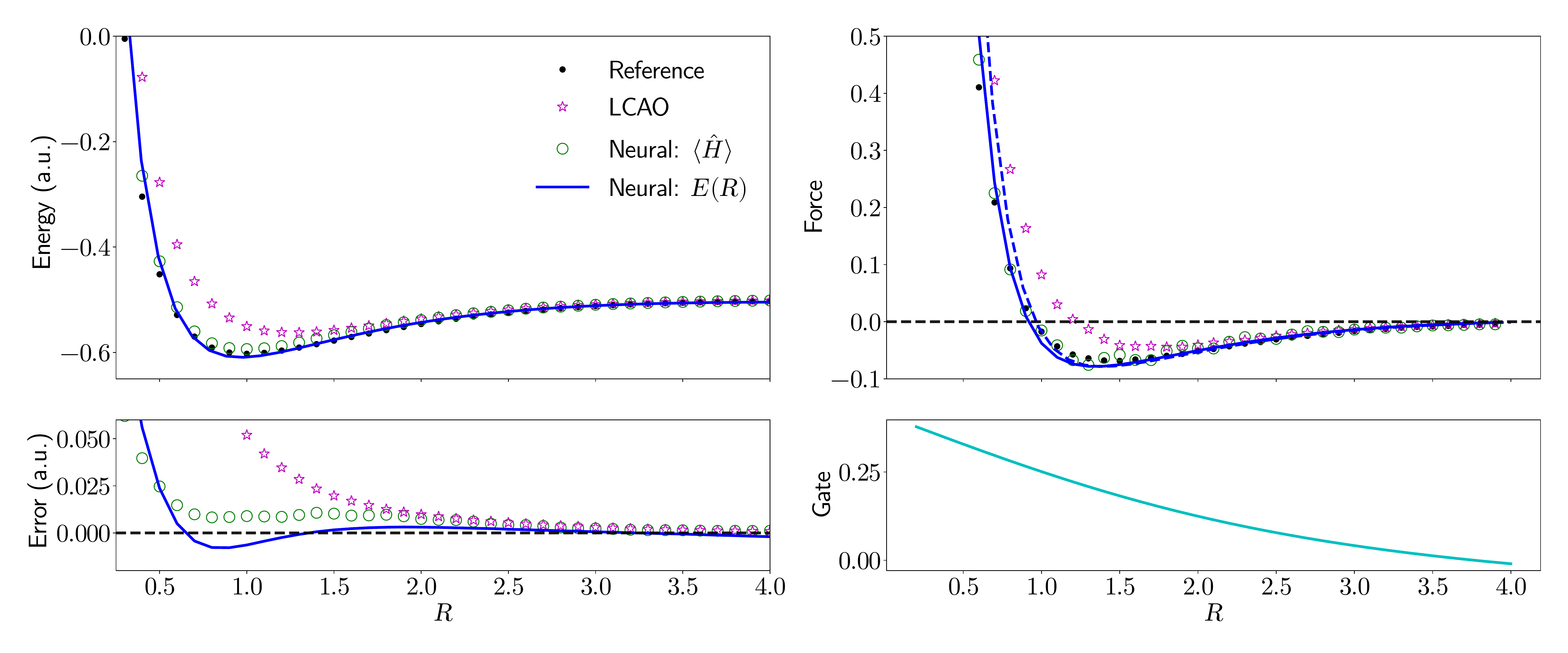}
  \caption{{\bf Left panels}: Energy and relative error in atomic units (a.u.), as functions of $R$. {\bf Right panels}: force, $dE/dR$, and gate function $f(R)$; finite differences are used to calculate the forces in all the cases except for the dashed blue line where autograd is used.}
  \label{fig:fig4}
\end{figure}

{\bf Neural  energy potential surfaces and wavefunction:}
Figures~\ref{fig:fig3} and \ref{fig:fig4} show the results after  training the NN. 
Figure~\ref{fig:fig3}(b) shows the neural $\psi$ (solid blue) and $\psiL$ (dashed red) along the $x$-axis for two different values of $R$ indicated by the dashed black lines. The two plots on the right show perspective 3D views of $\psi(x,y,0)$ for the same two values of $R$. 

In Fig.~\ref{fig:fig4} the upper left panel  shows the  total energy of the H$^+_2$ system, namely the sum of the electronic energy and the classical electrostatic energy of the nuclei  ($1/2R)$.  For the electronic energy we compare $E(R)$ coming directly from the PINN, the expectation value $\langle \HH \rangle$ calculated with both the neural $\psi(R)$ and $\psiL$, and a ground truth reference energy calculated in  Ref.~\cite{truthH2}.  The lower left panel shows the differences from the reference energy, where we observe that the error in $E(R)$ is sometimes negative, in violation of the variational principle, while $\langle\HH \rangle$ is never smaller than the ground truth. The lower right panel shows the gate function, which gives increased weight to the BU  for smaller $R$ as expected. 
%
The upper right panel of Fig. \ref{fig:fig4} shows the force between  the ions, obtained by differentiating the energy with respect to $R$. Since $E(R)$ is a continuous and  differentiable function of $R$, we use autograd to calculate the force (dashed blue line). The other energies are computed using finite differences. The force from $E(R)$ using finite differences (solid blue line) deviates from the autograd calculation when the slope of the force is large and thus numerical error is accumulated. 
This shows the  advantage of  using the proposed  $E(R)$ that eliminates errors from the  numerical derivatives.  

\vspace{-0.5cm} 
\section{Conclusions}
In this study we present a novel deep learning approach to obtain continuous and differentiable parametric  wavefunctions and potential energy surfaces for molecular systems,
using a PINN architecture for solving the first-principles quantum mechanical equations. 
We validate this approach by studying  the electronic ground state of H$^+_2$. 
Although we presented results only for a simple molecule, 
the proposed architecture is generalizable to other similar systems.
The most important advantage of this approach is the ability to obtain neural-net  wavefunctions and energies that are continuous and differentiable in the parameter space of the nuclear coordinates. Having such a form of the solutions is useful for  further computations like the calculation of forces and vibrational frequencies.  

\section*{Broader Impact}
 Solving eigenvalue partial differential equations is an important goal in many scientific fields including engineering, applied physics, and quantum chemistry. Solving these equations can be extremely demanding and frequently prohibitive due to the limitations of existing numerical methods. New technologies and more efficient methods for solving differential equations are crucial to accelerate progress in scientific research.
In this work, we introduced a  deep learning framework for solving eigenvalue partial differential equations for parametric potential functions. The proposed neural network is able to learn parametric eigenvalue and eigenfunction surfaces, namely neural solutions in terms of  the independent variables and of the modeling parameters.  We demonstrated the method's efficacy  by solving the stationary Schr\"odinger equation for the hydrogen molecular ion with one electron.
Generalization to more complicated problems is the subject of ongoing research.

 {\bf Societal and Environmental Impact: }
  We are not aware of any negative social impact from solving differential equations with neural networks, although, as for any scientific discovery, the results of using this mathematical tool depend on the intentions of the user. As far as the environment is concerned, the proposed method can be generalized to provide a wide range of solutions with a single training. Also, an approximated solution that is embedded in the neural network structure drastically reduces the training time and thus, the consumption of energy to arrive at the solution. For the specific problem presented in this work, the training of the NN took less than five minutes on an NVIDIA Tesla GPU. 

\section*{Acknowledgements}
Research was sponsored by the Army Research Office and
was accomplished under Cooperative Agreement Number W911NF-21-2-0147 and Grant Number
W911NF-21-1-0184; the STC Center for Integrated Quantum Materials, NSF Grant No.
DMR-1231319; and NSF DMREF Award No. 1922172.


\bibliographystyle{apalike2}
\bibliography{references}

\end{document}